# Automatic Recognition of Coal and Gangue based on Convolution Neural Network

HuichaoHong, LixinZheng, JianqingZhu, ShuwanPan, KaitingZhou

(1.Institute of Huaqiao University, Quanzhou, Fujian Province 362021, China;

2. Industrial Engineering Research Center, universities and intelligent systems in Fujian Province, Quanzhou, Fujian Province 362021, China)

**Abstract:** We designed a gangue sorting system, and built a convolutional neural network model based on AlexNet. Data enhancement and transfer learning are used to solve the problem which the convolution neural network has insufficient training data in the training stage. An object detection and region clipping algorithm is proposed to adjust the training image data to the optimum size. Compared with traditional neural network and SVM algorithm, this algorithm has higher recognition rate for coal and coal gangue, and provides important reference for identification and separation of coal and gangue.

Coal is mainly energy in the word which play a very important role in the development of the national economy. But there are plenty of coal gangue in addition to coal in the process of mining. Coal gangue is mainly composed of rock. Its density is large, ash content is high and calorific value is small. If coal gangue mixed in industrial production will seriously affect the quality and efficiency of combustion. So coal and coal gangue must be sorted before use. The traditional sorting is mainly used by manual selection and mechanical separation. Manual selection is not only costly but also seriously restricts the efficiency of mineral processing. Mechanical separation will cause serious pollution and loss of concentrated coal during the process of raw ore crushing.

In recent years, with the popularization of computer and high performance imaging equipment, the method of image based sorting is easy to be realized. Compared with coal gangue, coal gangue and black color, light color, its surface texture also have great differences. So the gray distribution and peak value of coal and gangue are also different. It can be seen that the image gray and texture analysis of coal and coal gangue will help to identify them.

In the past feature engineering methods, such as the extraction of feature texture based on the gray level symbiotic matrix, [1-3] is used to identify the coal gangue. Document [4] modeled the distribution of various coal components in the feature space by using the hybrid model, and used simple nearest neighbor decision rule to classify each pixel in the image.

Unlike the above algorithms, the convolution neural network is machine learning techniques which driven by big data set, greatly improved in the field of computer vision technology. Its texture features without artificial structure and can independently study coal and coal gangue, and based on the nonlinear model of training the original data into higher expression levels and the more abstract. For classification tasks, high-level expression can enhance the ability to distinguish input data and weaken the unrelated factors[5-6].

However, because the collected tags data set is smaller, it may lead to the training of convolutional neural network (CNN), and it is very prone to overfitting. In order to overcome this problem, the common method is to use the transfer of learning strategies, namely the use of large data sets in a pre-trained CNN model, using the weight of the CNN model as the initial set of related tasks or as a fixed feature extractor [7].

In this paper, we propose a transfer learning strategy, which uses the CNN features learned from other fields to initialize the convolution neural network.

The main contributions of this study are as follows:

1) According to our understanding, we first apply the convolution neural network to the classification of coal and coal gangue.

2) The strategy of classifying our data from the features that are learned from other data sets in the field is studied.

3) Because the proposed method has a better recognition rate compared with the traditional algorithm, it may be used as the sorting task of coal and coal gangue.

The rest of the paper is organized as follows: the first section briefly introduces the composition of the gangue sorting system. The second section explains the development of the convolution neural network, the components and the network models we adopted. The proposed method, the experiment setup respectively in the third section of the detailed description. Finally, summarize the whole article and e xplain the importance of the research.

## 1 System Components

The sorting system of coal and gangue is mainly composed of vibrating feeder, conveyor belt, industrial camera, industrial computer and separating mechanism. The structure of the system is shown in figure 1.

In the process of sorting, the mixture of coal and coal gangue is put into a vibrating feeder, which is separated as far as possible through a vibrating feeder, and then fed into a conveyor belt, which is captured by an industrial camera placed above the conveyor belt to collect the mixture of coal and coal gangue on the conveyor belt, and is determined by the computer, The coal and coal gangue are fed into the corresponding classification trough by separating system.

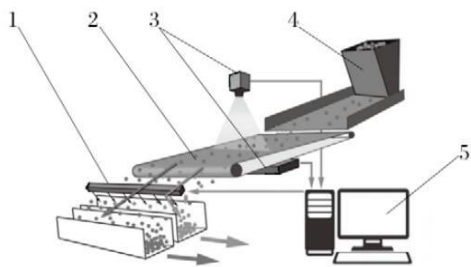

1. separation mechanism 2. conveyor belt 3. industrial camera 4. vibrating feeder 5. industrial computer

Figure 1. Structure sorting system

## 2 Convolutional Neural Network

Convolutional neural network is first proposed 1980 by Fukushima in 1980[8] ,and be further perfected byYannLecnn et al[9].In 2011 and 2012, Danciresan and et al have been further optimized, enabling CNN to achieve excellent results in multiple image databases MNIST, NORB, HWDB and CIFAR10.CNN network structure is closer to biological neural network has a unique advantage in speech recognition and image processing, in particular to obtain a good visual image processing application. In this study, the structure is selected CNN AlexNet [10], it has achieved great success in image classification.

Although more excellent network models have been put forward since 2012, the usual practice is to deepen the number of networks. However, it usually means that processing a graph requires more computing time, which makes the system's real-time performance unguaranteed. This is also the reason for this paper to choose AlexNet as a network model, and to a certain extent, it takes into account the accuracy and speed.

Convolution layer is an important component of the convolution neural network. There are two important attributes in the convolution layer: local connection and weight sharing. In a locally connected network, each node is only partially connected to the position of the upper node. This feature has many characteristics, such as reducing parameter, not easy to fit, saving memory space and training and prediction time. The output of the convolution layer is represented by equation(1).

$$h_{i,j} = f\left(\sum_{m}^{M}\sum_{n}^{N} w_{m,n} x_{i+m, j+n} + b_{m,n}\right) \qquad (1)$$

Where x represents the input feature image, i and j denote input feature image element index,w is the convolution kernel of M x N, m and n represent convolutional kernel parameter index, B represents bias, h represents output neuron.

Relu is used as the activation function in AlexNet, compared with the traditional sigmoid activation function of large amount of calculation, prone to gradient disappeared and other shortcomings, the Relu function will make the output part of a neuron is 0, so the network is sparse, and reduces the mutual dependency relation parameters, alleviate the overfitting problem. The computational efficiency by using the maximum pool turn to improve, through the selection of maximum value of local image regions as the area after pooling value. Another technique used to speed up training and avoid overfitting is a method of setting out the output of hidden neurons at random by a certain probability(drop out),which reduces complex and common adaptation of neurons, and compels each neuron to have more information.

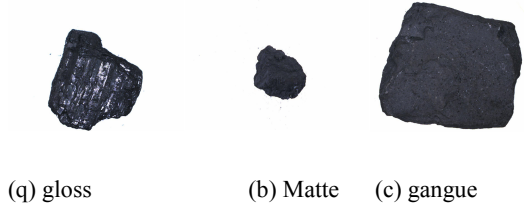

(q) gloss  (b) Matte  (c) gangue

Figure 2. samples

AlexNet also used the LRN(Local Response Normalization) layer. In 2012, Hinton et al put forward the LRN side to imitate the side inhibition mechanism of the biological nervous system, and create a competition mechanism for the activities of the local neurons, so that the larger response value is relatively larger, and the generalization ability of the model is improved. The method of calculation is shown in equation (2):

$$h'_{i,j} = h_{i,j}/(1+(\alpha/R)\sum_i\sum_j h^2_{i',j'})^\beta \quad (2)$$

Where R represents the local size, i and j represents a local neighboring neuron index, α represents zoom parameters, β represents exponential.

In Alexnet, each layer of weight and bias parameters are updated by minimizing the softmax loss function by the random gradient descent method, as shown in equation (3):

$$J(W) = -\sum_s\sum_p \frac{\{y^{(s)}=p\}log(W^{(p)T}x^{(s)})}{\sum_q q^{exp(W^{(q)T}x^{(s)})}} \quad (3)$$

Where s represents the index of a sample in a batch, and p and q are the index of the class label. Its weight updating equation (4) is shown as follows:

$$w(n) = w(n-1) - \eta \times (\frac{\partial J(w)}{\partial w(n)} + \alpha \Delta w(n-1))$$
$$n = 1, 2, 3...$$

Where $\Delta w(n)$ represents the increments of weight increment, $\alpha$ represents the momentum factor, $\lambda$ represent the regularization coefficient.

## 3 Experimental results and analysis

### A. Data pre-processing

Coal and coal gangue samples from Shanxi, including 2012 images(769 glossy coal, 660luster ,583 coal gangue).

In this experiment, 1521 images were selected for training, while the rest were used for testing

Table 1, the experimental data set coal and gangue

| data | Types of | Quantity |
|---|---|---|
| Training Set | Gangue | 417 |
|  | Matt (coal) | 514 |
|  | Gloss(coal) | 590 |
| Test Set | Gangue | 166 |
|  | Matt (coal) | 146 |
|  | Gloss(coal) | 179 |

In the process of obtaining the image of coal and coal gangue, the size of the broken pieces will be different, as shown in Figure 2. For small pieces, it means only a small piece of feature recognition, while other areas are blank. Therefore, we propose an algorithm to detect the coal and coal gangue areas and keep the coal image area as much as possible. The specific practice, as shown in Table 1.

Table 2. object detection region clipping algorithm

| |
|---|
| Step 1: Using the Sobel operator to calculate the gradient in the direction of X and Y* |
| Step 2: subtracting the gradient in the direction of the Y in the direction of the X, leaving the image area with a high level gradient and a low vertical gradient.* |
| Step 3: using Gauss filter to remove the noise in the image and smooth the high frequency noise in the image* |
| Step 4: using the maximum inter class difference method for image two value.* |
| Step 5: looking for image contour, and compressing horizontal direction, vertical direction, diagonal direction elements, retain only the end coordinates of the direction, so as to get 4 points as rectangular contour. |
| Step 6: get a rectangle with the smallest area of the point set. |
| Step 7: through step 6, a rectangular area with 4 vertex information can be obtained. Find the maximum |
| minimum of the X and Y coordinates of the four vertices. The height of the new image equal to maxY-minY and width equal to maxX-minX. |
| Note: the steps above with * can find the corresponding function prototype in OpenCV. |

After the processing of the above algorithm, the result is shown in Figure3.

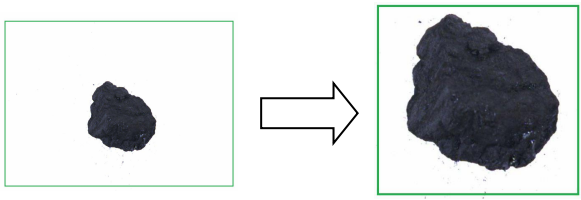

Figure 3. effect of object detection and region cutting algorithm

At the same time, when we train the network, we need to use a lot of images, so we enhance the data of training set, and the training data set is 5 times of the original data training dataset.
- We extracted from each of the five positions of the original image: the four corners and center.
- We rotate the origin image to make each image generate another 5 images.
- Adding Gaussian random noise for some origin image.

Because the cropped image can be any shape or size, and the need for a fixed input dimension in the full connection layer ,so we adjust the processed image to 256 * 256 by linear interpolation algorithm.

In order to obtain better classification results, we subtract the statistical mean of the data from each sample and carry out the normalization process. Above procedure is is shown in Figure 4.

### B. How to Fine-Tune the Network

Many works point that the first few convolutional layers tend to learn features that resemble edges, lines, corners, shapes, and colors, independent of the training data. More specifically, earlier layers of the network contain generic features that should be useful to many tasks. Since we can define and characterize front as edges, lines, and corners. The task is to train a CNN model to extract generic features that can be useful to our work. To train a CNN model, we need a large data set. However, our data set was not large enough to train the full CNNs; therefore, fine-tuning becomes the preferred option to extract the features.

Fine-tuning a network is a procedure based on the concept of transfer learning. Specifically, it is a process that adapts an already learned model to a novel classification model. There are two possible approaches of performing fine-tuning in a pretrained network: first is to fine-tune all the layers of the CNNs. The second approach is to keep some of the earlier layers fixed (to avoid overfitting) and fine-tune only the higher level layers of the network. In the first approach, the classifier layer is removed from the pretrained CNNs and the rest of the CNNs are treated as a fixed feature extractor. In the second approach, the initial layers are frozen to keep the generic features already learned and the final layers are adjusted for the specific task. In other words, fine-tuning uses the parameters learned from a previous pre-trained network on a specific data set and then adjusts the parameters from the current state for the new data set. In this paper, we fine-tuned all layers except fully connected layer and assumed that the features from all layers were important for our task.

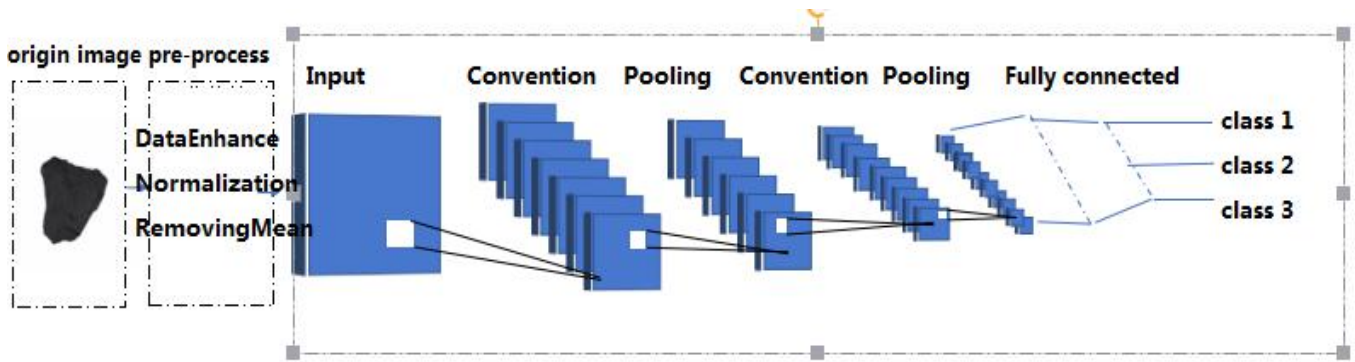

Figure 4. Recognition process of coal and coal gangue

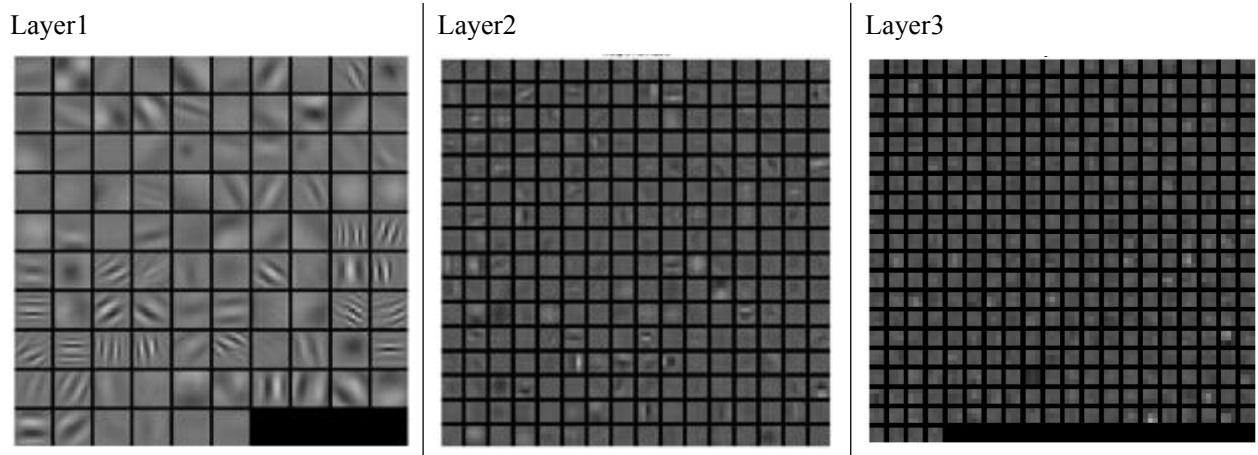

Figure 4 (a). Visualized effect of the convolution kernel of the pre training model obtained in the ImageNet image database training

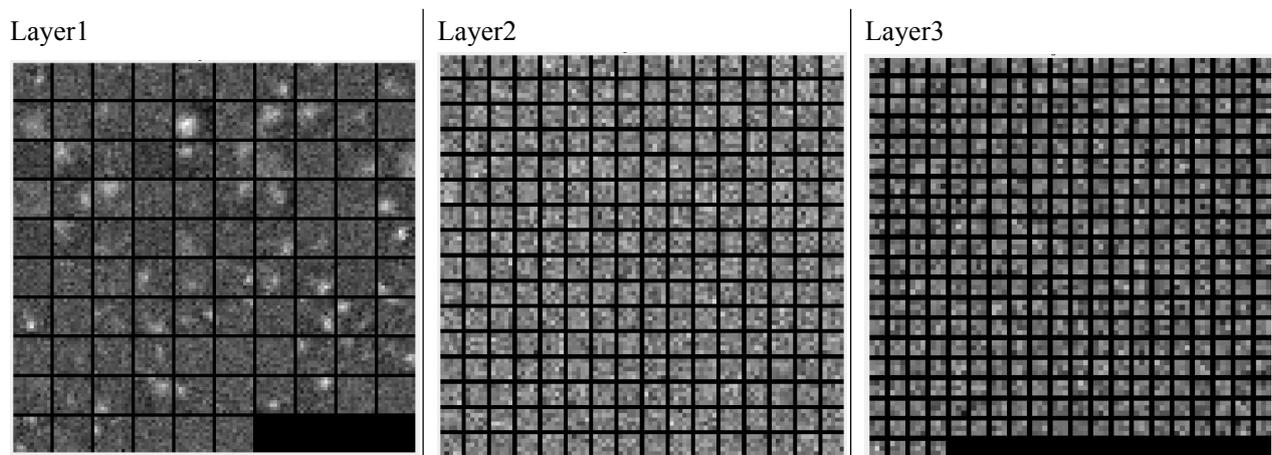

Figure 4 (b). Visual effect of convolution kernel without pre training mode

Figure 4. AlexNet convolution kernel visualization

### C. Experiment

In order to adapt to the classification task of this experiment, we will adjust the output layer's class quantity by 3.

During training, computers need memory space to store parameters such as weights and bias. The setting of batch size (Batch-Size) also depends on computer hardware parameters. In this experiment, the graphics card we used is NVIDIA GTX1050 (Ti) which has 4GB of memory.In order to maximize the memory usage,we set the size of Batch-Size to 256.

Because prerequisites for fine-tuning is the weight of the pre -training model is meaningfu.When the learning rate is too large, it will be updated quickly, and it destroys the original weight information structure.

We use the momentum coefficient to speed up the training and ensure the stability of the training. From the perspective of intuition, if the current gradient direction is the same as the gradient direction of the previous step, we will increase the update of this step's weight and reduce the update if it's different.Therefore, the higher the setting of $\alpha$, the faster convergence of the network can be made. The lower $\alpha$ values are usually used in early iterations.We chose $\alpha$ equal to 0.9 since we are using the pre-training model in ImageNet whose weights are available for most classification tasks.

As shown in Figure 4, we visualize weights(layer1 to layer3) using pre training model and the weights without the pre training model. Figure 4 (b) has a lot of noise, figure 4 (a) is relatively smooth. The emergence of this situation, often means that the process of training data model caused by our lack of training.

Figure 5 shows that the fine-tuning method is used to train the network (2000 iteration), and the loss function curve is also smoother and faster. It only applies to the Gauss initialization method. The loss function curve is slower and the curve is more concussion.

In summary, our paper realized GLCM(the energy, contrast, entropy and correlation are extracted from the grayscale symbiotic matrix of the sample) + SVM algorithm, and compared with convolution neural network.

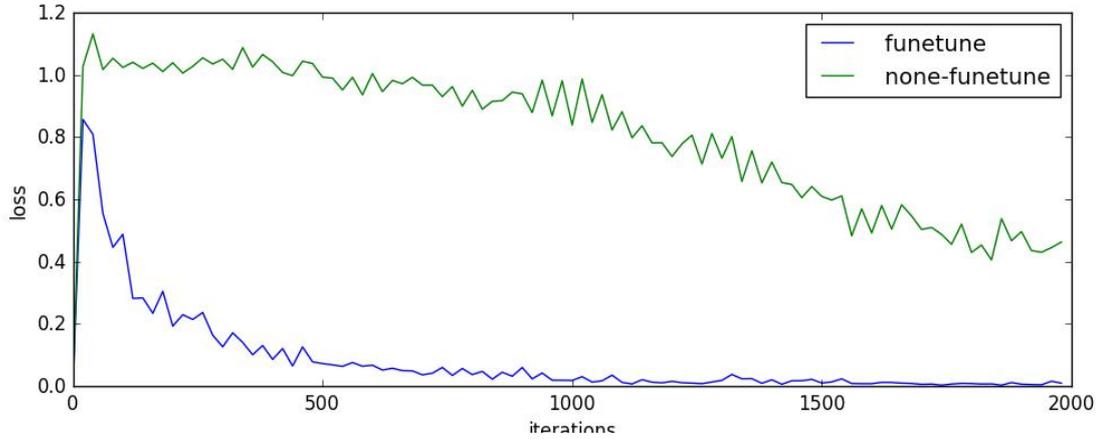

Figure5.The change of the Loss value of the 2000 iteration loss function

### 3.3 Evaluation Methods

As far as we know, the early commonly used image gray histogram statistical results as features, and then the gray level co-occurrence matrix is proposed, and the characteristics of coal and coal gangue is more conducive to the region, which is currently the most commonly used coal texture analysis method.

On the classifier, the BP neural network is usually used for classification in the early stage, which can realize nonlinear mapping, have self-learning ability, and have the ability to generalize. However, the lack of gradient descent method is slow and possible to enter the local minimum and training failure. Then, because of the emergence of SVM, it has overcome many shortcomings of BP neural network to a certain extent and is widely welcomed.

Value training fails, the newly added sample impact may occur due to learning or had learning. Then the advent of SVM, to a certain extent, to overcome the shortcomings of BP neural network and most popular.

For the three classification tasks, the recognition rate (RecG), the recall (Rec) and the accuracy (Pre) are used as the performance indicators to measure the algorithm:

$$\text{Recg} = \frac{\sum_i m_{ii}}{\sum_i \sum_j m_{ij}} \qquad (9)$$

$$\text{Rec}_j = \frac{\sum_i m_{ii}}{\sum_i m_{ji}} \qquad (10)$$

$$\text{Pre}_j = \frac{\sum_i m_{jj}}{\sum_j m_{ji}} \qquad (11)$$

Where i and j represent different categories.

The results of the experiment are shown in Table 4.Comparing the results of GLCM+SVM with our method, we can see that our proposed method provides better results than the above, and achieves 96.6% in three categories. In all the schemes, the best results have been achieved.

Table 3. algorithm performance comparison

| method | Recg | Rec_G | Rec_C1 | Rec_C2 | Pre_G | Pre_C1 | Pre_C2 |
| --- | --- | --- | --- | --- | --- | --- | --- |
| CNN | 0.94 | 0.90 | 0.92 | 0.96 | 0.95 | 0.88 | 0.92 |
| Finetune- CNN | 0.96 | 0.93 | 0.94 | 0.98 | 0.98 | 0.91 | 0.95 |
| SVM + GLCM | 0.89 | 0.84 | 0.96 | 0.92 | 0.96 | 0.91 | 0.82 |

## 4 Conclusion

In recent years, the demand for visual - based object recognition has increased dramatically, and it can improve efficiency and protect the environment. Taking coal and coal gangue recognition as the application background, first of all, after preprocessing the images acquired in a certain order, we study the method of accurate identification using convolution neural network. But at the same time, it is also noted that the type and quantity of experimental samples taken are limited, so the experimental results of the authors are not necessarily applicable to all coal and gangue identification. Increasing the number of sample images and the types of sample sources will improve the accuracy and universality of automatic recognition, which will also be one of the future directions of the author.

### References

[1] Liang H, Cheng H, Ma T, et al. Identification of Coal and Gangue by Self-Organizing Competitive Neural Network and SVM[C]// International Conference on Intelligent Human-Machine Systems and Cybernetics. IEEE, 2010:41-45.

[2] .Min H E, Wang P P, Jiang H H. Recognition of coal and stone based on SVM and texture[J]. Computer Engineering & Design, 2012, 33(3):1117-1121.

[3] Tan C, Yang J. Research on extraction of image gray information and texture features of coal and gangue image[J]. Industry & Mine Automation, 2017.

[4] P. A. van Vuuren et al., "Using visual texture analysis to classify raw coal components," 2015 International Conference on Systems, Signals and Image Processing (IWSSIP), London, 2015, pp. 212-215.

[5] Gao Z Y, Wang A, Liu Y, Zhang L, Xia Y W. Intelligent Fresh-tea-leaves Sorting System Based on Convolution Neural Network [J].Transactions of the Chinese Society for Agricultural Machinery,2017,(07):1-12.

[6] WANGKun, LIUDamao, Intelligent identification for tea state based on deeplearning[J]. Journal of Chongqing University of Technology, 2015,12(29):120-126.

[7] A.Krizhevsky, I. Sutskever, G. E. Hinton, "Imagenet classification with deep convolutional neural networks", Adv. Neural Inf. Process. Syst., pp. 1106-1114, 2012.

[8] Fukushima K. Neocognitron: A self-organizing neural network model for a mechanism of pattern recognition unaffected by shift in position[J]. Biological Cybernetics, 1980, 36(4):193-202.

[9] Lecun Y, Jackel L D, Cortes C, et al. Learning Algorithms For Classification: A Comparison On Handwritten Digit Recognition[J]. Neural Networks the Statistical Mechanics Perspective, 1995:261--276.

[10] Krizhevsky A, Sutskever I, Hinton G E. ImageNet classification with deep convolutional neural networks[C]// International Conference on Neural Information Processing Systems. Curran Associates Inc. 2012:1097-1105.

[11] L. Haonan, S. Baojin, H. Yaqun, H. Jingfeng and H. Qiongqiong, "Research on identification of coal and waste rock based on GLCM and BP neural network," 2010 2nd International Conference on Signal Processing Systems, Dalian, 2010, pp. V2-275-V2-278.